\documentclass[runningheads]{llncs}
\usepackage[T1]{fontenc}
\usepackage{graphicx}
\usepackage{booktabs} 
\begin{document}
\title{Local Language Models for Context-Aware Adaptive Anonymization of Sensitive Text}
%
%
\author{Aisvarya Adeseye\inst{1}\orcidID{0009-0003-2401-3076} \and
Jouni Isoaho\inst{1}\orcidID{0000-0002-5789-3992} \and
Seppo Virtanen\inst{1}\orcidID{0000-0002-9487-3018}
\and
Mohammad Tahir\inst{1}\orcidID{0000-0002-6273-4603}}
\authorrunning{A. Adeseye et al.}
%
\institute{Department of Computing, University of Turku, Turku, Finland \\
\email{\{aisvarya.a.adeseye, jouni.isoaho, seppo.virtanen, tahir.mohammad\}@utu.fi}
}
\maketitle              
\begin{abstract}
Qualitative research often contains personal, contextual, and organizational details that pose privacy risks if not handled appropriately. Manual anonymization is time-consuming, inconsistent, and frequently omits critical identifiers. Existing automated tools tend to rely on pattern matching or fixed rules, which fail to capture context and may alter the meaning of the data. This study uses local Large Language Models (LLMs) to build a reliable, repeatable, and context-aware anonymization process for detecting and anonymizing sensitive data in qualitative transcripts. We introduce a Structured Framework for Adaptive Anonymizer (SFAA) that includes three steps: detection, classification, and adaptive anonymization. The SFAA incorporates four anonymization strategies: rule-based substitution, context-aware rewriting, generalization, and suppression. These strategies are applied based on the identifier type and the risk level. The identifiers handled by the SFAA are guided by major international privacy and research ethics standards, including the GDPR, HIPAA and National Standards TCPS 2, and OECD guidelines. This study followed a dual-method evaluation that combined manual and LLM-assisted processing. Two case studies were used to support the evaluation. The first includes 82 face-to-face interviews on gamification in organizations. The second involves 93 machine-led interviews using an AI-powered interviewer to test LLM awareness and workplace privacy. Two local models, LLaMA and Phi were used to evaluate the performance of the proposed framework. The results indicate that the LLMs found more sensitive data than a human reviewer. Phi outperformed LLaMA in finding sensitive data, but made slightly more errors. Phi was able to find over 91\% of the sensitive data and 94.8\% kept the same sentiment as the original text, which means it was very accurate, hence, it does not affect the analysis of the qualitative data.

\end{abstract}
\section{Introduction}

Qualitative research can provide valuable insights by collecting narratives from real-life situations \cite{ref_lncs1}. However, these narratives often contain personal, organizational, and location-specific information that can pose privacy risks \cite{ref_lncs2}. Anonymization is important in academic and applied fields because failure to do it will lead to a breach of ethical obligations, privacy regulations, and a lack of trust in the research process \cite{ref_lncs3}. Moreover, manual anonymization remains the most dominant method. However, they are time-consuming, labor-intensive, and error-prone. Although automated tools that use pattern or static rules exist, nevertheless, these tools are inflexible despite being fast, they also struggle to understand human language nuances \cite{ref_lncs11}, and may often distort the original meaning when rewriting sensitive information.

Human understanding of context and machines' ability to process information vary, which creates a problem of how to create an anonymization method that is both fast and accurate. Recent advances in large Language models (LLMs) create an opportunity to bridge this gap. Consequently, locally hosted LLMs can be customized to understand information context \cite{adeseye2025}, preserve data sovereignty without relying on external cloud services.

To address this problem, this study develops a Structured Framework for an Adaptive Anonymizer (SFAA) using local LLMs to detect, classify, and adaptively anonymize sensitive information in qualitative texts.  The goal is to protect privacy and ensure compliance with the data protection standards. We define the following research questions to achieve the aim of this study.
\begin{itemize}
        \item \textbf{RQ1: }How accurately does a local LLM detect and classify sensitive information compared to independent expert manual analysis?

        \item \textbf{RQ2: }How effective are the different adaptive anonymization strategies for privacy protection while preserving meaning and maintaining the integrity of qualitative analysis results?
\end{itemize}

This study makes two key contributions to the literature. First, it presents a precise and repeatable three-step anonymization method tested with human and LLM-based approaches. Second, it proves that LLMs can find and anonymize sensitive text while maintaining its meaning, reducing the need for time-consuming manual work.

\section{Literature Review}
Anonymization is the process of removing or altering personal details in data so that individuals cannot be identified \cite{ref_lncs4}. In qualitative research, certain parts of information could reveal someone's identity, such as names, locations, job roles, or unique stories. The goal is to protect people's privacy while still keeping the data useful and meaningful \cite{ref_lncs14}. However, anonymization is not pseudonymization. Pseudonymization replaces identifiers, but does not fully prevent re-identification in some cases \cite{ref_lncs15}. 

True anonymization should prevent re-identification. It supports ethical research practice that meets legal and regulatory requirements, such as the General Data Protection Regulation (GDPR) in Europe, The Health Insurance Portability and Accountability Act (HIPAA) in the United States, and the Australian National Statements on Ethical Conduct in Human Research (National Statement) in Australia \cite{scheibner2020healthdata}. These and other standards provide the foundation for the collection, storage, and sharing of data. If anonymization is done incorrectly, researchers could face serious ethical and legal problems.

Several tools exist that can help with anonymization, such as Dedoose, NVivo, and Atlas.ti. These tools support manual coding and offer some features for tagging or hiding identifiers  \cite{ref_lncs6}. Some other automated tools that use pattern-matching rules to remove sensitive terms like names, addresses, or emails \cite{ref_lncs7} also exist, for example MITRE Identification Scrubber Toolkit (MIST) and ARX Data Anonymization Tool.  Furthermore, more advanced tools like Presidio by Microsoft use natural language processing (NLP) to find and mask personal details \cite{ref_lncs8}. Consequently, these tools work well for direct identifiers but miss indirect or context-based information and may sometimes change the meaning of the text, which makes it difficult to utilize them for data analysis. For example, a phrase like "the only woman who is an IT manager in our department" does not mention a specific name, but still clearly indicates a particular person because of the context. Also, most existing tools are not designed to handle qualitative data in its raw format, which is important for understanding people's thoughts and feelings. Interview transcripts are more complex as their meaning depends on the context, tone, and flow of language. These tools are usually made for structured formats like spreadsheets or forms. 

Consequently, many researchers prefer to do this manually \cite{ref_lncs12}. Manual methods can be more accurate when dealing with complex but short qualitative data. However, these methods are time-consuming, inconsistent, and depend on the skill of the analyst performing the task \cite{ref_lncs12}. Also, this method is unsustainable when the data is large or when working under strict deadlines. 

Current literature indicates that there is no reliable, scalable, and context-aware method available for qualitative data anonymization. Large language models like ChatGPT and Gemini offer potential in mitigating this gap. However, these commercial cloud-based systems raise privacy and ethical concerns \cite{adeseye2025ichms}. Additionally, local LLM alternatives like LLaMA and Phi are unexplored. Therefore, this study proposes a structured framework for local LLMs to detect, classify, and anonymize sensitive information in qualitative transcripts. The proposed approach provides a private and context-aware approach to anonymization, filling a critical gap in the literature.

\section{Methodology}

\begin{figure}
    \centering
    \includegraphics[width=0.8\linewidth]{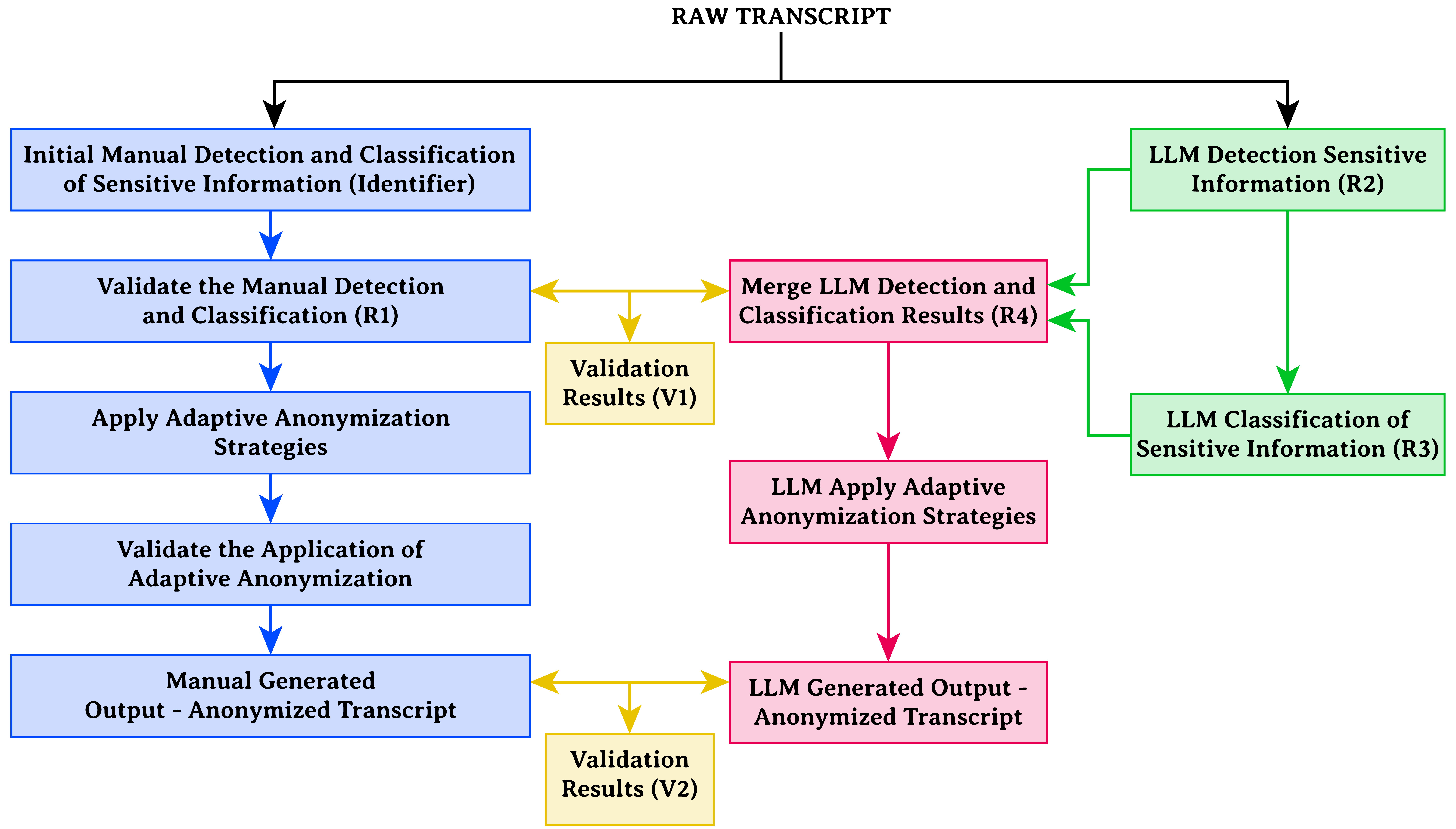}
    \caption{Hybrid Anonymization Workflow Integrating Manual Review and LLM-Assisted Processing for Sensitive Qualitative Data.}
    \label{fig:workflow}
\end{figure}

When anonymizing transcripts, it is important to strike a balance between context sensitivity and processing efficiency. Therefore, we adopted a hybrid anonymization workflow (Figure \ref{fig:workflow}) that incorporates manual coding and also LLM-assisted coding approaches. Manual annotation can identify indirect but strong details specific to certain cultures, people, or a place, which an LLM might overlook. Contrastingly, LLMs are quick when processing large qualitative data. Therefore, two stages of validation to check and ensure that the LLM and manual output created preserved accurately the meaning of the transcript were created.

\subsection{Datasets}

Qualitative data from two different case studies were utilized to test the proposed SFAA. The first was a face-to-face interview of 82 participants about the challenges of introducing gamification for workforce studies. Each interview lasted for 45 to 60 minutes and generated qualitative transcripts of between 8,000 and 13,000 words. The second was an AI-powered interview of 93 participants about how LLMs are used at their workplace. The transcript length ranged from 4000 to 11000 words.

Using both case studies to test the SFAA has several advantages. They provide different types of interview transcripts; the first was a human-led interview, while the second was an AI-led interview using a local LLM. This difference will help test how well the anonymization process works with different communication and expression mechanisms. The differences in content, length, and how the interviews were done ensure that the framework is tested in a realistic and diverse way.

\subsection{Experiment Setup}

We used two language models, LLaMA v3.2 (3B parameters) and Phi v3.2 (4B parameters), on a computer with an 8-core processor, 32 GB of memory, and a 500 GB hard drive running offline and locally. Smaller models (3B \& 4B) were chosen because they are fast, resource-efficient, cheap, and work well. However, larger models are resource-intensive and expensive to run.

\section{Structured Framework for Adaptive Anonymizer (SFAA)}

\begin{figure}
    \centering
    \includegraphics[width=1\linewidth]{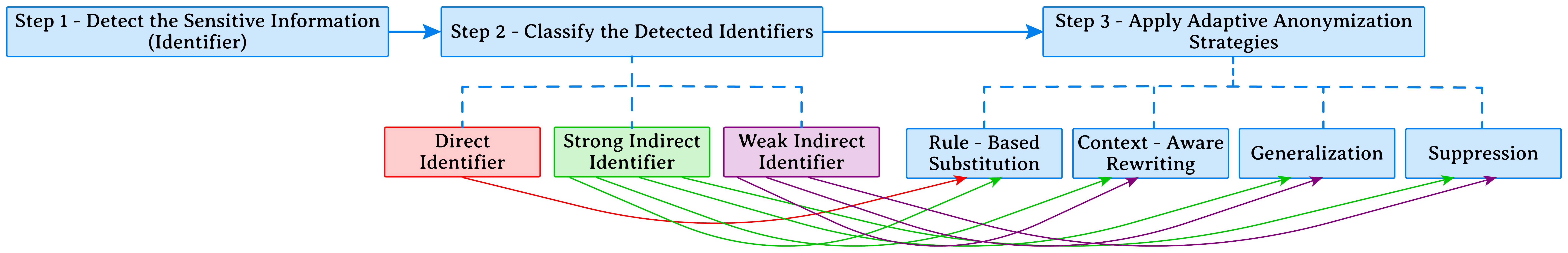}
    \caption{Structured Framework for Adaptive Anonymizer (SFAA): A Three-Step Process for Identifying, Classifying, and Anonymizing Sensitive Information in Qualitative Transcripts.}
    \label{fig:overview}
\end{figure}

The Structured Framework for Adaptive Anonymizer (SFAA) comprises three major steps such as detection, classification, and adaptive anonymization as seen in figure \ref{fig:overview}. Firstly, sensitive information from the transcripts is found by the system to ensure that the risks are identified early. Secondly, the identified information is categorized into three types: direct, strong indirect, and weak indirect identifiers, making it possible to understand how sensitive the data is and what type of protection is needed. Third, the appropriate adaptive anonymization strategy is applied to the classified data. These strategies include substitution, context-aware rewriting, generalizations, and suppression. Each step is important: detection ensures that all sensitive content is found, classification guides the level of protection needed, and adaptive anonymization ensures that the final output is both useful and private. 

\subsection{Step 1 - Identification}
\begin{figure}
    \centering
    \includegraphics[width=1\linewidth]{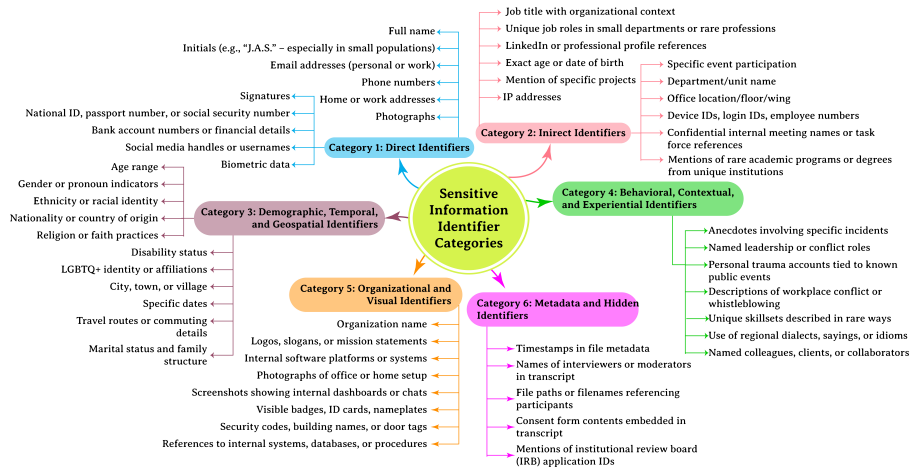}
    \caption{Comprehensive Identifier Categories for Detecting Sensitive Information in Qualitative Transcripts}
    \label{fig:step1}
\end{figure}

Step 1 focuses on detecting sensitive information in qualitative transcripts by identifying different types of personal, contextual, and organizational data, referred to as identifiers. The list is intentionally extensive to cover a wide range of details that could lead to the re-identification of individuals. The identifiers are grouped into six main categories: \textit{direct identifiers},  \textit{indirect identifiers},    \textit{behavioral, contextual, and experiential identifiers},  \textit{organizational and visual identifiers},  \textit{metadata and hidden identifiers}, and \textit{demographic, temporal, and geospatial identifiers}, as shown in Figure \ref{fig:step1}. These categories were selected because they reflect the most common privacy risks found in global qualitative research contexts. Each group targets a specific aspect of identifiability, from clearly stated personal details to subtle behavioral cues or embedded file metadata.

The development of this framework was guided by a review of major international privacy and research ethics standards. These include the Health Insurance Portability and Accountability Act (HIPAA) in the United States, the General Data Protection Regulation (GDPR) in the European Union, the Tri-Council Policy Statement: Ethical Conduct for Research Involving Humans (TCPS 2) in Canada, the Australian National Statement on Ethical Conduct in Human Research, and OECD Guidelines on the Protection of Privacy and Transborder Flows of Personal Data. These documents share common principles of minimizing harm, respecting confidentiality, and ensuring data protection during the research lifecycle.

Practical insights from over 170 qualitative interview data were utilized to refine the identifier categories in the list. A simple yes or no tagging method was adopted by the researchers (manually created) and the locally hosted LLMs. This provided insights on how humans and machines detect sensitive information. This step is important because it ensures that no sensitive information is omitted before classification and anonymization.

\subsection{Step 2 - Classification}

\begin{figure}
    \centering
    \includegraphics[width=0.7\linewidth]{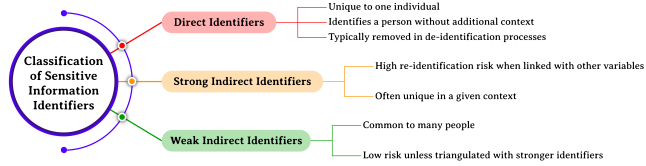}
    \caption{Three-Level Classification of Sensitive Information Identifiers Based on Re-identification Risk: Direct, Strong Indirect, and Weak Indirect Categories to Guide Privacy-Preserving Strategies.}
    \label{fig:step2}
\end{figure}

Identified sensitive information in the transcripts was classified into three main groups: direct, strong indirect, and weak indirect identifiers as seen in the figure \ref{fig:step2}.

This helps to determine the risk associated with each group and the anonymization strategy to adopt. Direct identifiers are details that clearly identify an individual, such as names or ID numbers. Strong indirect identifiers are not unique on their own, but can still identify someone when combined with other information, like job titles or small-group roles. Weak indirect identifiers, like general locations or age pose a lower risk but can still identify someone if combined with other stronger clues. These three categories were developed by reviewing existing privacy guidelines and research practices, and refined by analyzing real-world transcript data. We kept the classification system simple with only three levels to make it easier to apply it consistently across different datasets, while still capturing the full range of re-identification risks.

\subsection{Step 3 - Adaptive Anonymization}

\begin{figure}
    \centering
    \includegraphics[width=1\linewidth]{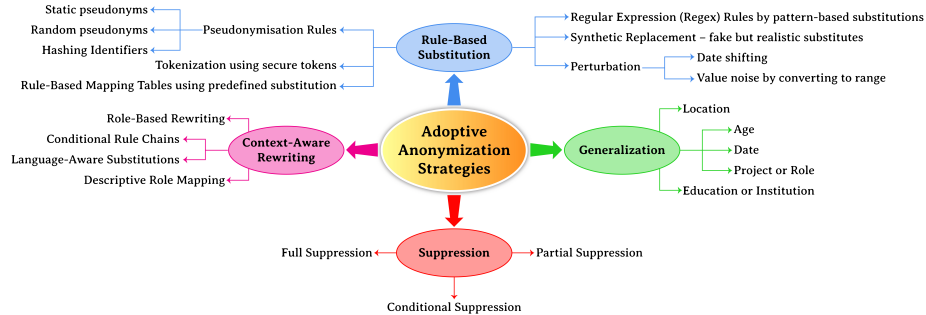}
    \caption{Overview of Adaptive Anonymization Strategies: A Four-Part Framework Including Rule-Based Substitution, Context-Aware Rewriting, Generalization, and Suppression for Protecting Sensitive Information in Qualitative Data.}
    \label{fig:step3}
\end{figure}

In Step 3, we apply adaptive anonymization strategies to the sensitive information identified and classified in the previous stages as seen in Figure \ref{fig:step3}.


After determining the type and risk level of each identifier, we transform, generalize, or suppress the data in a way that protects privacy without compromising the meaning or analytical value of the transcript. We used four main strategies: rule-based substitution, context-aware rewriting, generalization, and suppression. Every strategy targets different identifier categories and risk levels. Figure \ref{fig:step3} shows all four strategies and the techniques adopted for each one of them.  This structured approach maintains consistency and ensures flexibility. Also, it supports both machine and human-assisted anonymization workflows. A more detailed explanation of each strategy for the framework is provided below.

\textbf{Rule-Based Substitution} is a technique used to replace sensitive information with safe alternatives via structured and predictable methods like pseudonyms; the use of static, random aliases, or hashing identifiers to replace original terms. Tokenization is another method; sensitive terms are replaced with secured but reversible tokens. Also, rule-based mapping uses a predefined substitution table to ensure replacement uniformity across various datasets. Additionally, regular expression rules, which find and replace patterns is another method. Moreover, synthetic replacement involves the insertion of fake but realistic values. Perturbation includes methods such as shifting of dates or converting values to a given range. They are effective for replacing strong indirect and direct identifiers, such as names or institutions with consistent placeholders. These approaches are sufficient for standardizing anonymization without compromising data quality; they were chosen because they cover the major rule-based substitution method and balance privacy with repeatability and structure, especially for large datasets.

\textbf{Context-Aware Rewriting} is used when rule-based substitution is not enough to maintain the meaning or flow of the transcript. This method applies more intelligent rewriting techniques based on the context in which an identifier appears. Role-based rewriting adjusts references based on a person's function rather than their identity. This helps to preserve the original meaning of the text. Conditional rule chains apply substitutions only when specific conditions are met, which helps to reduce over-redaction. This ensures that sensitive information is removed without losing the original context. Language-aware substitutions preserve the tone, grammar, and natural phrasing of the text. This helps to maintain the readability of the anonymized data. Also, descriptive role mapping replaces identifiers with safe narrative labels that still hold analytical meaning. Context-aware rewriting can be used to rewrite both strong and weak indirect identifiers. It helps prevent semantic distortion, a risk associated with using rule-based substitution alone. The methods discussed balance readability and privacy; there was no need to include other complex natural language generation methods.

\textbf{Generalization} technique prevents re-identification by reducing details while at the same time keeping data useful. Details such as exact location, age, date, project, role, or educational institution are replaced with ranges. This is useful for weak and some strong indirect identifiers; it preserves patterns while concealing exact values by using consistent placeholders. 

\textbf{Suppression} technique is used to remove information that cannot be anonymized safely by applying full, partial, or conditional suppression. Full suppression means removing the entire information, partial suppression means removing only a part of it, while conditional suppression means the information is only removed when certain thresholds are breached. This method is utilized for weak and strong indirect identifiers where replacement or rewriting is not practicable. Suppression is a last resort measure in this framework, but could be essential when no safer alternative exists that can preserve ethical standards of not exposing sensitive information.

\section{Application and Validation of SFAA}
This section presents in greater detail how the three phases of the framework, such as identification, classification, and anonymization were applied across both case studies using manual and LLM-based methods.
\subsection{Identification}
\begin{figure}
    \centering
    \includegraphics[width=0.8\linewidth]{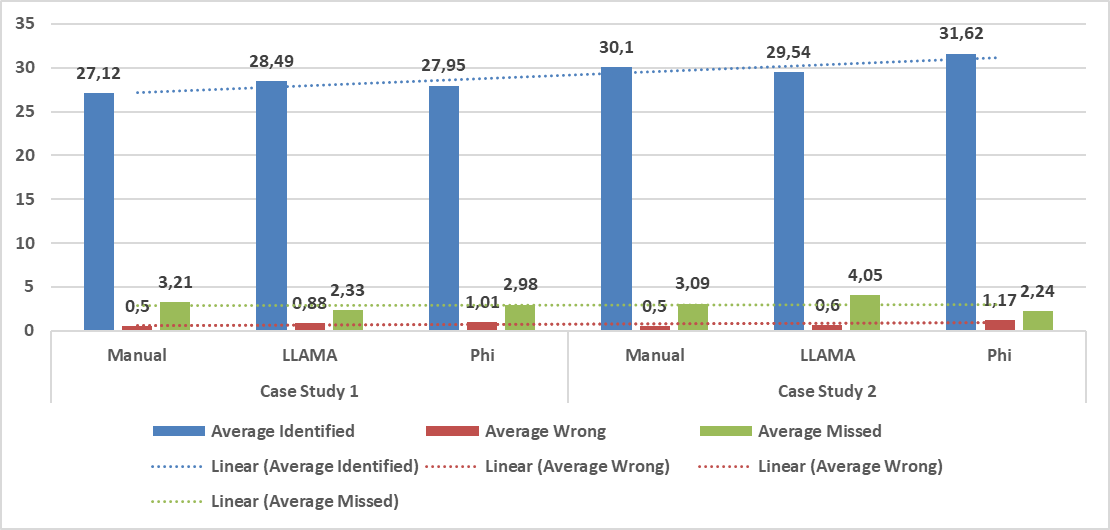}
    \caption{Average Number of Identified, Wrongly Tagged, and Missed Sensitive Identifiers by Method Across Two Case Studies}
    \label{fig:compare1-1}
\end{figure}

10 transcript samples were drawn from each of the two case studies as a representative sample of each case study, and the results scaled proportionally across all 82 and 93 transcripts in both case studies. The results include manual evaluation and the two LLM models (LLaMA \& Phi) as seen in Figure \ref{fig:compare1-1}.

\textbf{Performance Summary for Case Study 1 - } The results indicated that the LLAMA and Phi outperform manual annotation by identifying more sensitive items per transcript while also missing fewer sensitive items. Phi was slightly more sensitive; hence, it had a higher hallucination rate than LLAMA. The manual method had the least wrongly identified sensitive items and also the highest number of missed identification.

\textbf{Performance Summary for Case Study 2 - } The results indicated that Phi outperforms manual annotation and LLaMA by identifying more sensitive items per transcript while also missing fewer sensitive items. Phi remained the most sensitive as it had the highest hallucination rate again. The manual method remained the only method that had the least wrongly identified sensitive, while LLaMA had the highest missed identification.

\begin{figure}
    \centering
    \includegraphics[width=0.7\linewidth]{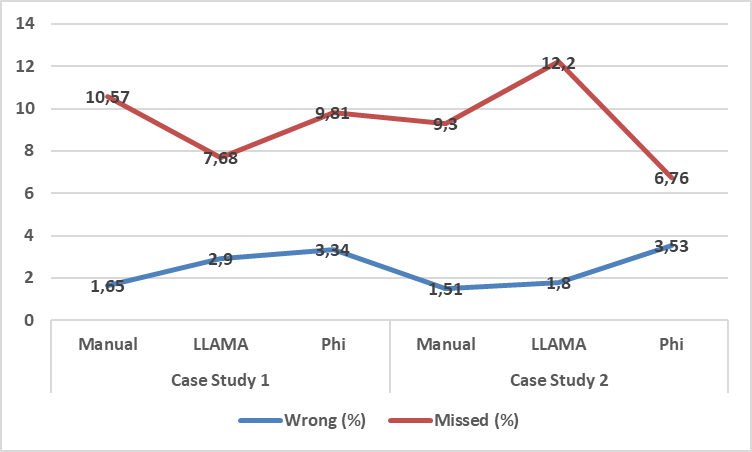}
    \caption{Comparison of Missed and Wrong (Hallucinated) Identifier Rates (\%) Across Manual, LLaMA, and Phi Methods in Two Case Studies. The percentages are computed by comparing the average number of items found in Figure \ref{fig:compare1-1} to the total number of items for that category}
    \label{fig:compare1-2}
\end{figure}

\textbf{Cross-Case Comparison Summary - } In both case studies, the LLMs (Phi \& LLaMA) identified more sensitive items than the manual method, which means they show strong potential for anonymization automation. However, both LLMs had higher wrongly identified sensitive items (hallucination) when compared to the manual method, which had the lowest wrongly identified sensitive item percentage, as seen in \ref{fig:compare1-2}, indicating that it remained the best method for mitigating against a wrong detection. Also, Phi had the lowest missed sensitive items percentage across both case studies. Overall, combining human insights with LLM detection capability offers the best method to identify data privacy issues when collecting qualitative data.

\subsection{Classification}

\begin{figure}
    \centering
    \includegraphics[width=0.8\linewidth]{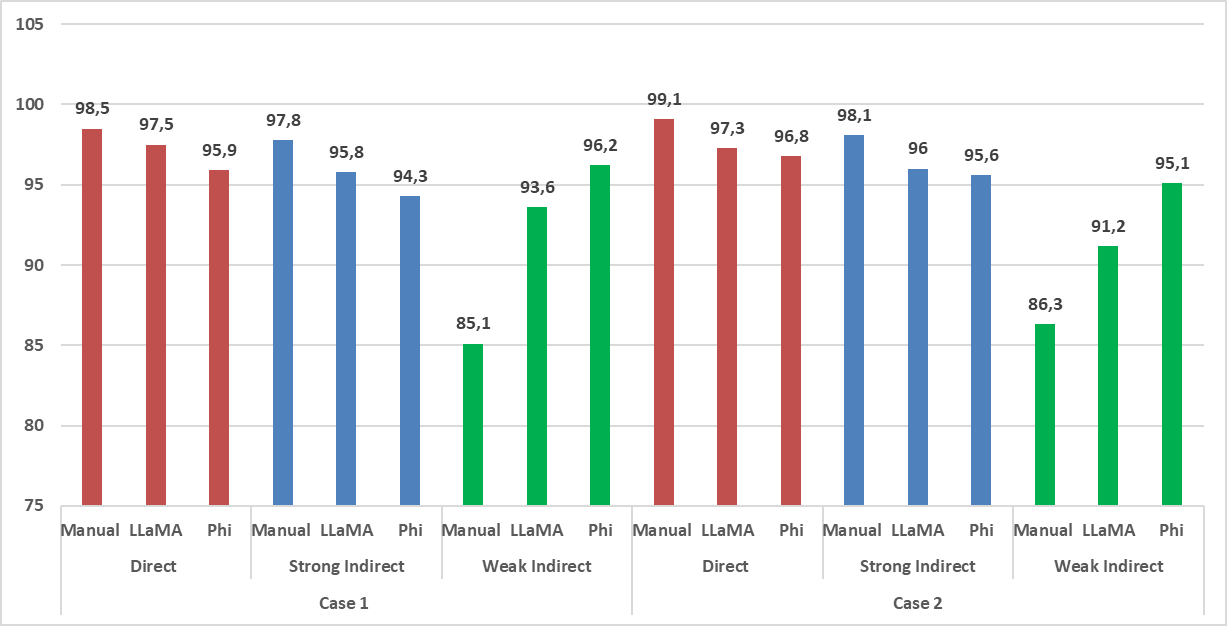}
    \caption{Classification Accuracy of Direct, Strong Indirect, and Weak Indirect Identifiers Across Manual, LLaMA, and Phi Methods in Case Studies 1 and 2}
    \label{fig:compare2-1}
\end{figure}

The accuracy of classification across both case studies remains consistent. As seen in Figure \ref{fig:compare2-1}, Phi excelled at consistently classifying weak indirect identifiers with a percentage of 96.2\% and 95.1\% for both case studies. LLaMA performed relatively close as well, far better than the manual method, which struggled in this regard, indicating that LLMs excel at identifying subtle clues that humans overlook. Contrastingly, the manual method was best at classifying direct identifiers with 98.5\% and 99.1\% accuracy for both case studies. Similarly, it was also the best way to classify strong indirect identifiers. However, the performance of both LLMs in classifying direct and strong indirect identifiers was not far off.

\begin{figure}
    \centering
    \includegraphics[width=0.8\linewidth]{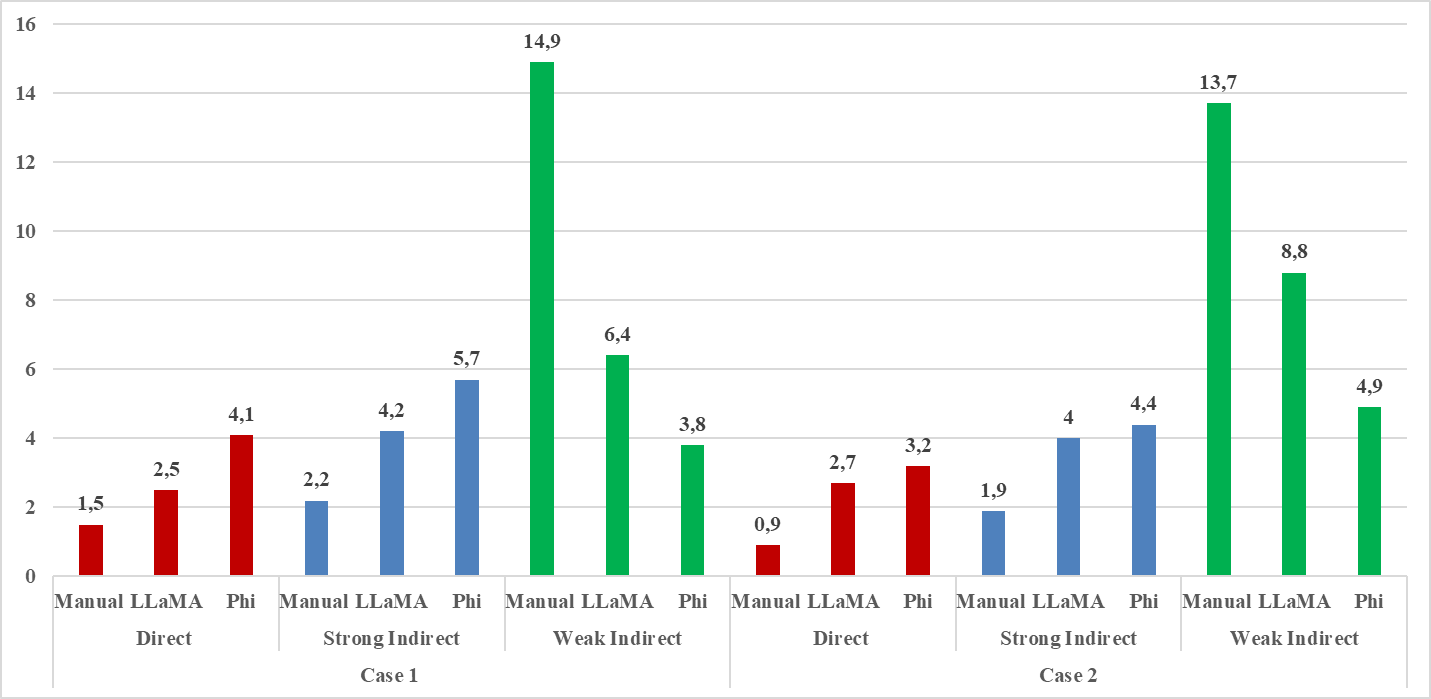}
    \caption{Classification Error Rates (\%) for Direct, Strong Indirect, and Weak Indirect Identifiers Using Manual, LLaMA, and Phi Methods Across Case Studies 1 and 2}
    \label{fig:compare2-2}
\end{figure}

Also, Figure \ref{fig:compare2-2} shows the number of errors committed by the manual and both LLMs. In both case studies, the manual method wrongly categorized weak identifiers with a percentage of 14\% and 13\% for both case studies, with a big disparity in comparison to both LLMs; Phi made the fewest mistakes, indicating the importance of LLMs to help detect these weak indirect identifiers. However, the manual method had the lowest error percentage at 1.5\% and 0.9\% for direct identifiers. Similarly, it had the lowest error percentage for strong indirect identifiers as well, but the performance of both LLMs where not far off.

These results emphasize the importance of integrating LLMs into the anonymization workflow, particularly for improving the detection and classification of weak indirect identifiers that the manual method struggles to classify. When validated and paired with human insight, the classification capabilities of LLMs can significantly enhance consistency and depth of qualitative data anonymization.

\subsection{Adaptive Anonymization}

The final step of the anonymization process involves applying the four strategies across both case studies: rule-based substitution, context-aware rewriting, generalization, and suppression. Each strategy is used to address different types of sensitive information, depending on the context and risk level.

Rule-based substitution was used for strong indirect and direct identifiers such as names, company titles, and institutional affiliations. This strategy replaces these identifiers with consistent and repeatable rules. Context-aware rewriting was used to preserve the meaning of identifiers that are embedded in descriptive or role-based statements. This is important when rigid replacements could distort the narrative. Generalization was applied when exact details were not needed for the analysis like job titles, location, and others; these were replaced with broader terms. 
Usernames and ID number was considered high risk; hence, they were suppressed because they could not be rewritten safely. Table~\ref{tab:anonymization_examples_combined} contains some examples of how the four strategies were applied to some transcript content.

\begin{table}[h]
\centering
 
\caption{Examples of Adaptive Anonymization Strategies Across Both Case Studies}
\label{tab:anonymization_examples_combined}
\resizebox{\textwidth}{!}{ 
\begin{tabular}{|p{5.2cm}|p{2.5cm}|p{5.2cm}|}
\hline
\textbf{Original Text} & \textbf{Anonymization Strategy} & \textbf{Output} \\
\hline
``My name is Rajeev and I introduced gamification at OptiCore.'' & Rule-Based Substitution & ``My name is [Person\_1] and I introduced gamification at [Company\_1].'' \\
\hline
``I'm Dr. Nilmini from the Computer Science department at University of Kelaniya.'' & Rule-Based Substitution & ``I'm [Doctor] from the tech department at [University\_1].'' \\
\hline
``Only I and the senior developer were allowed to test the prototype game.'' & Context-Aware Rewriting & ``A limited number of staff were selected to test the pilot system.'' \\
\hline
``Since I was leading the LLM compliance training, I had to attend legal briefings.'' & Context-Aware Rewriting & ``As the training lead on responsible AI use, I had to attend privacy briefings.'' \\
\hline
``I work at the regional branch in Jaffna, managing 15 team members.'' & Generalization & ``I work at a northern regional branch in Sri Lanka, managing a medium-sized team.'' \\
\hline
``The AI course is offered only through the Department of Data Ethics.'' & Generalization & ``The AI course is offered through a specialized academic department.'' \\
\hline
``My staff username is NLMAdmin24, and I accessed GPT-4 three times last week.'' & Suppression & ``[Redacted], and I accessed a large language model three times last week.'' \\
\hline
``The internal gamification project is tagged under file ID 22HR-918-MT.'' & Suppression & ``The internal gamification project is tagged under [Redacted].'' \\
\hline
\end{tabular}
}
\end{table}

The combination of all four strategies was integral in handling this sensitive data. Local LLMs played vital roles in anonymizing both direct and indirect identifiers; they were helpful for context evaluation and selecting the right method for each case study and were most effective for context-aware rewriting, improving consistency across transcripts.

 \section{Discussion}

An evaluation of the performance of local LLMs for adaptive anonymization was conducted. Three major metrics were utilized for the evaluation: accuracy, recall, and precision. Accuracy measures the number of correct classifications. However, recall measures how many correct sensitive elements were detected, while precision measures correct detections \cite{ref_lncs9}. These metrics were computed against a manually validated reference set that provided a benchmark for comparing the results. 

In Case Study 1, Phi achieved high scores: 91.3\% for recall, 96.6\% for precision, and 94.2\% for accuracy. While for Case Study 2, the scores were 93.5\% for recall, 95.1\% for precision, and 94.9\% for accuracy. LLaMA performed slightly worse when compared to Phi. Manual anonymization had high precision (above 98\%), but lower recall and accuracy due to missed indirect identifiers.

Each anonymization strategy had its own strengths. Rule-based substitution achieved the highest precision (99.2\%), but was limited to direct and some strong indirect identifiers. Context-aware rewriting offered strong recall (94.5\%) and balanced accuracy with the original textual meaning. Generalization performed well for roles and locations, with all metrics exceeding 92\%. Suppression achieved near perfect precision, but at the expense of data loss.

We also evaluated the impact of anonymization on qualitative analysis. Thematic coding remained highly consistent, with over 92\% agreement before and after anonymization. Frequency analysis showed minimal change. Word counts and topic clusters varied by less than 3\%. Sentiment and impact analysis were more sensitive, with the overuse of rule-based or suppression strategies distorting sentiment. Context-aware rewriting preserved emotional meaning more effectively, achieving a 94.8\% average alignment score.

Overall, LLMs offered faster and more consistent anonymization with improved detection of subtle identifiers. When supported by a structured framework, they can help balance privacy protection and analytical integrity more effectively than manual methods.

\section{Conclusion}
This study presented a structured framework for adaptive anonymization using local LLMs. The goal of the framework is to ensure privacy while preserving the analytical meaning of the qualitative data. For the two case studies evaluated, local LLMs, especially Phi achieved a high recall and strong precision of over 91\% and 95\% respectively, outperforming manual methods in detecting subtle identifiers. Rule-based substitution had the highest precision while context-aware rewriting preserved meaning in complex transcripts. Suppression was used selectively to minimize information loss. 

The output of the anonymization process did not have a significant impact on downstream analysis. Themes, frequency, and sentiment alignment remained consistent at over 94\% similarity before and after anonymization. However, some limitation exists; LLM performance may vary when the language or dialect is changed, or for non-interview-based data. Also, hallucination from LLM outputs required careful data filtering to elicit only useful information from the LLMs. For future work, we will explore multilingual anonymization, test datasets from other diverse disciplines like healthcare and law.
Also, LLM prompting strategies could be further refined and optimized to reduce hallucination. Generally, the framework offers a scalable, privacy-preserving anonymization solution for qualitative research.

%
%
%

\end{document}